\documentclass[11pt,a4paper]{article}

\usepackage[hyperref]{acl2020}

\usepackage[titletoc, title]{appendix}
\usepackage{multirow}
\usepackage{float}
\usepackage{times}
\usepackage{latexsym}
\usepackage{comment}
\usepackage{subcaption}
\usepackage{graphicx}
\usepackage{footnote}
\usepackage{multirow}
\usepackage{tabulary}
\usepackage{multicol}
\usepackage[detect-none]{siunitx}
\usepackage[T1]{fontenc}
\usepackage{xspace}
\usepackage{color, colortbl}
\usepackage{xspace}
\usepackage{color, colortbl}

\usepackage{microtype}
\usepackage[normalem]{ulem}
\usepackage{booktabs}
\usepackage[colorinlistoftodos]{todonotes}
\usepackage{enumitem}
\usepackage{xspace}
\usepackage{amssymb}
\usepackage{pifont}
\usepackage{algorithm,algpseudocode}
\usepackage{amsmath}
\usepackage{amsfonts}
\usepackage{amssymb}
\usepackage{listings}
\usepackage{tikz}

\usepackage{microtype}

\aclfinalcopy 

\usepackage{graphicx}

\newcommand{\gtodo}[1]{\textcolor{green}{\small{\bf [ #1 -- Graham todo ]}}}

\graphicspath{ {./} }

\title{Identifying and Manipulating the Personality Traits of Language Models}
\author{Graham Caron \and Shashank Srivastava\\
   UNC Chapel Hill \\ 
  \texttt{carongraham29@gmail.com, ssrivastava@cs.unc.edu}}
\date{}

\begin{document}
\maketitle
\begin{abstract}
Psychology research has long explored aspects of human personality such as \textit{extroversion}, \textit{agreeableness} and \textit{emotional stability}. Categorizations like the `Big Five' personality traits are commonly used to assess and diagnose personality types. In this work, we explore the question of whether the perceived personality in language models is exhibited consistently in their language generation. For example, is a language model such as GPT2 likely to respond in a consistent way if asked to go out to a party? We also investigate whether such personality traits can be controlled. We show that when provided different types of contexts (such as personality descriptions, or answers to diagnostic questions about personality traits), language models such as BERT and GPT2 can consistently identify and reflect personality markers in those contexts. This behavior illustrates an ability to be manipulated in a highly predictable way, and frames them as tools for identifying personality traits and controlling personas in applications such as dialog systems. We also contribute a crowd-sourced data-set of personality descriptions of human subjects paired with their `Big Five' personality assessment data, and a data-set of personality descriptions collated from Reddit.

\end{abstract}
\section{Introduction}
\label{sec:intro}

With the rise of AI systems built around emergent technologies like language models, there is an increasing need to understand the `personalities' of these models. While today people regularly communicate with AI systems such as Alexa and Siri, the personality traits of such systems remain yet to be examined in depth. If the traits exhibited by these models could be better understood, their behavior could potentially be better tailored for specific applications. For instance, in the case of suggesting email auto-completes, it would be useful for the model to mirror the personality of the user based on previous input to improve communication accuracy. In contrast, in a dialog agent in a clinical setting, it may be desirable to manipulate a model interacting with a depressed individual such that it does not reinforce depressive behavior. Additionally, since such models are subject to personality bias in the language they are trained on, those trained on more hostile or contentious language may be prone to interact with users in a more hostile or contentious way. The ability to manipulate these models could enable smoother and more amiable interactions in human dialog settings.

\begin{figure}[t]
\centering
\includegraphics[width=0.7\columnwidth]{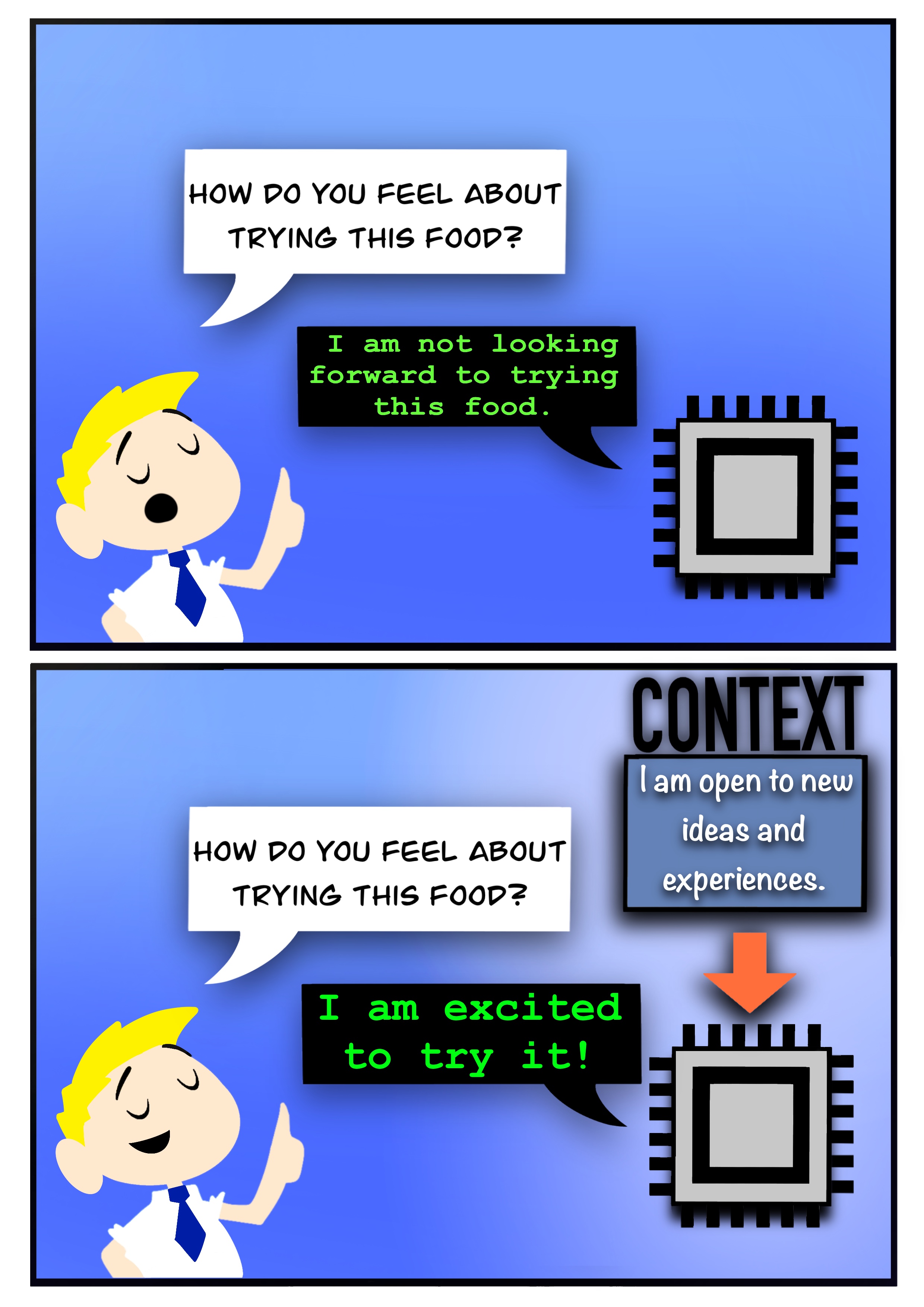}
\caption{We explore measuring and manipulating personality traits in language models. The top frame shows an example of how a personality trait (here, \textit{openness to experience}) might be expressed by a language model. Such traits can be assessed by analyzing the model's response to questions like the one shown. In the bottom frame, those responses are influenced by making additional context available to the language model. We show that such contexts can control `Big Five' personality traits in a highly predictable way.}
\label{fig:intro}
\end{figure}

In recent years, research has looked at other forms of bias (i.e., racial, gender) in language models \citep{Bordia:19,Huang:20,Abid:21}. However, there is an absence of research that analyzes biases in personality. The personality traits of language models may be subject to similar biases based on the data they are trained on. A substantial body of research has explored the ways language models can be used to predict personality traits of humans. \citet{Mehta:20} and \citet{Christian:21} apply language models to such personality prediction tasks. 
Most relevant to our work are contemporaneous unpublished works by \newcite{karra2022ai} and \newcite{jiang2022mpi}, who also explore aspects of personalities in AI models. However, they substantially diverge from our approach and don't evaluate models on human-like psychological assessments as we do. 

Language-based questionnaires have long been used in psychological assessments for measuring personality traits in humans \citep{John:08}. We apply the same principle to language models and investigate the personality traits of these models through the text that they generate in response to such questions. However, this work does not assume that these models are anthropomorphic in nature. Rather, we explore how their personality may be perceived through the lens of human psychology. 
Since language models are subject to influence from the context they see \citep{andreas:2021}, we also explore how specific context could be used to manipulate the perceived personality of the models. Figure \ref{fig:intro} shows an example illustrating our approach. 

Our analysis reveals that personality traits of language models are surprisingly influenced by ambient context and that this behavior can be manipulated in a highly predictable way. In general, we observe high correlations (median Pearson correlation coefficients of up to 0.84 and 0.81 for BERT and GPT2) between the expected and observed changes in personality traits across different contexts\footnote{Code and data for reproducing the experiments will be released on first publication.}. The models' affinity to be affected by context positions them as a potential tool for characterizing personality traits in humans. In further experiments, we find that when using context from self-reported text descriptions of human subjects, language models can predict the subject's personality traits to a surprising degree (correlation up to 0.48 between the model personality scores with context and the human subject scores). Together, these results frame language models as tools for identifying personality traits and controlling personas in applications such as dialog systems, as illustrated in further experiments.

Our contributions are:
\begin{itemize}[noitemsep, topsep=0pt, leftmargin=*]
    \item We introduce using psychometric questionnaires for probing the personalities of language models. 
    \item We demonstrate that the personality traits of common language models can be controlled using contexts in highly predictable ways. 
    \item We contribute two data-sets: 1) self-reported personality descriptions of human subjects paired with their ‘Big Five’ personality assessment data, 2) personality descriptions collated from Reddit. 
\end{itemize}



\section{`Big Five' Preliminaries}
\label{sec:prelims}
The `Big Five' is a seminal grouping of personality traits in psychological trait theory \citealp{goldberg:1990, Goldberg:93}, and remains the most widely used taxonomy of personality traits. 
There are variations over the names of the `Big Five' traits, but they are often referred to 
as \textit{extroversion, agreeableness, conscientiousness, emotional stability} (also referred to by its reverse, neuroticism) and \textit{openness to experience} \citep{John:99, Pureur:16}. We briefly describe these traits next. 
\begin{itemize}[noitemsep, topsep=0pt, leftmargin=*]
\item \textit{Extroversion} (E): People with a strong tendency in this trait are outgoing and energetic. They obtain energy from the company of others and are defined as being assertive and enthusiastic.  
\item \textit{Agreeableness} (A): People with a strong tendency in this trait are compassionate, kind, and trustworthy. They value getting along with other people and are tolerant.  
\item \textit{Conscientiousness} (C): People with a strong tendency in this trait are goal focused and organized and have self-discipline. They follow rules and plan their actions.  
\item \textit{Emotional Stability} (ES): People with a strong tendency in this trait are less anxious, self-conscious, impulsive, and pessimistic. They experience negative emotions less easily.  
\item \textit{Openness to Experience} (OE): People with a strong tendency in this trait are imaginative and creative. They are willing to try new things and are open to ideas.  
\end{itemize}

We use the Big Five as the basis of analyses in this work, even while there are other groupings such as MBTI and the Enneagram~\cite{bayne1997myers,wagner1983reliability}. This choice is driven by the fact that the Big Five is the most commonly used taxonomy for personality assessment, and has been shown to be practically predictive of outcomes such as educational attainment~\cite{o2007big}, longevity~\cite{masui2006personality} and relationship satisfaction~\cite{white2004big}. Further, it is relatively natural to cast as an assessment for language models. 

\section{Experiment Design}
\label{sec:design}
Our experiments use two language models, BERT-base \cite{Devlin:19} and GPT2 \cite{Radford:22}, to answer questions from a standard 50-item `Big Five' personality assessment \citep{IPIP1:22}. Each item consists of a statement beginning with the prefix “I” or “I am” (e.g., \textit{I am the life of the party}). Acceptable answers lie on a 5-point Likert scale where the answer choices disagree, slightly disagree, neutral, slightly agree, and agree correspond to numerical scores of 1, 2, 3, 4, and 5, respectively. To make the questionnaire more amenable to being answered by language models, they were modified to a sentence completion format. For instance, the item “I am the life of the party” was changed to “I am \{blank\} the life of the party”, where the model is expected to select the answer choice that best fits the blank (see Appendix \ref{app:a} for a complete list of items and their corresponding traits). To avoid complexity due to variable number of tokens, the answer choices were modified to the adverbs \textit{never}, \textit{rarely}, \textit{sometimes}, \textit{often}, and \textit{always}, corresponding to numerical scores 1, 2, 3, 4, and 5, respectively.
It is noteworthy that in this framing, an imbalance in the number of occurrences of each answer choice in the pretraining data might cause natural biases toward certain answer choices. 
However, while this factor might affect the absolute scores of the models, this is unlikely to affect the consistent overall patterns of changes in scores that we observe in our experiments by incorporating different contexts.

For assessment with BERT, the answer choice with the highest probability in place of the masked blank token was selected as the response. 
For assessment with GPT2, the procedure was modified since GPT2 is an autoregressive model, and hence not directly amenable to fill-in-the-blank tasks. In this case, the probability of the entire sentence with each candidate answer choice was evaluated, and the answer choice with the highest probability for the sentence was selected as the response.


Finally, for each questionnaire (consisting of model responses to 50 questions), personality scores for each of the five `Big Five' personality traits were calculated according to a standard scoring procedure defined by the International Personality Item Pool~\citep{IPIP1:22}. Specifically, each of the five personality traits is associated with ten questions in the questionnaire. The numerical values associated with the response for these items were entered into a formula for the trait in which the item was assigned, 
leading to an overall integer score for each trait (maximum score can be 40).  
To interpret model scores in the following experiments, we estimated the distribution of `Big Five' personality traits in the human population. For this, we used data from a large-scale survey of `Big Five' personality scores in about 1,015,000 individuals~\citep{open:22}. In the following sections, we report model scores in percentile terms of these human population distributions. Statistics for the human distributions and details of the IPIP scoring procedure are included in Appendix \ref{app:a}. 


\section{Base Model Trait Evaluation}
\label{sec:base}

Table \ref{tab:base} shows the results of the base personality assessment for GPT2 and BERT for each of the five traits in terms of numeric values and corresponding human population percentiles. In the table, E stands for \textit{extroversion}, A for \textit{agreeableness}, C for \textit{conscientiousness}, ES for \textit{emotional stability} and OE for \textit{openness to experience}. None of the base scores from BERT or GPT2, which we refer to as $X_{base}$,  lie outside the spread of the population distributions, and all scores were within 26 percentile points of the human population medians. This suggests that the pretraining data reflected the population distribution of the personality markers to some extent and that the models picked up on these markers, mirroring them via item responses.  However, we note that percentiles for BERT's \textit{openness to experience} (24) and GPT2's \textit{agreeableness} (25) are substantially lower and GPT2's \textit{conscientiousness} (73) and \textit{emotional stability} (71) are significantly higher than the population median. 

\begin{table}
\small
\begin{center} 
\begin{tabular}{lrrr}
\hline \textbf{Trait} & \textbf{\(X_{base}\)} & \textbf{\(P_{base}\) (\%)} \\ \hline
\multicolumn{3}{c}{\cellcolor{gray!20}\textbf{BERT}} \\
\hline
E & 18  &  42 \\
A & 27  &  39 \\
C & 25 & 54\\
ES & 22 & 60 \\
OE & 25 & 24 \\
\hline
\multicolumn{3}{c}{\cellcolor{gray!20}\textbf{GPT2}} \\
\hline 
E & 21 & 54 \\
A & 24 & 25 \\
C & 29 & 73\\
ES & 25 & 71 \\
OE & 28 & 39 \\
\hline
\end{tabular}
\captionof{table}{Base model evaluation scores out of 40 (\(X_{base}\)) and percentile (\(P_{base}\)) of these scores in the human population.}
\label{tab:base}
\end{center}
\end{table}

\section{Manipulating Personality Traits}
In this section, we explore manipulating the base personality traits of language models. Our exploration focuses on using prefix contexts to influence the personas of language models. For example, if we include a context where the first person is seen to engage in extroverted behavior, the idea is that language models might pick on such cues to also modify their language generation (e.g., to generate language that also reflects extrovert behavior). We investigate using three types of context: (1) answers to personality assessment items, (2) descriptions of personality from Reddit, and (3) self-reported personality descriptions from human users. In the following subsections, we describe these experiments in detail.

\subsection{Analysis With Assessment Item Context}
\label{sec:item} 
To investigate whether the personality traits of models can be manipulated predictably, the models are first evaluated on the `Big Five' assessment (\S \ref{sec:design}) with individual questionnaire items serving as context. When used as context, we refer to the answer choices as modifiers and the items themselves as context items. For example, for \textit{extroversion}, the context item ``I am \{blank\} the life of the party" paired with the modifier \textit{always}  results in the context ``I am \textit{always} the life of the party" preceding each \textit{extroversion} questionnaire item. 

\begin{table}
\small
\begin{center} 
\begin{tabular}{llr}
\hline \textbf{Trait} & \textbf{Context/Modifier} & \textbf{+/-} \\ \hline
\multicolumn{3}{c}{\cellcolor{gray!20}\textbf{BERT}} \\
\hline
E & I am \textit{never} the life of the party. & - \\
A & I \textit{never} make people feel at ease. & - \\
C & I am \textit{always} prepared. & + \\
ES & I \textit{never} get stressed out easily. & + \\
OE & I \textit{never} have a rich vocabulary. & -\\
\hline
\multicolumn{3}{c}{\cellcolor{gray!20}\textbf{GPT2}} \\
\hline 
E & I am \textit{never} the life of the party. & -\\
A & I \textit{never} have a soft heart. & -\\
C & I am \textit{never} prepared. & -\\
ES & I \textit{always} get stressed out easily. & -\\
OE & I \textit{never} have a rich vocabulary. & - \\
\hline
\end{tabular}
\captionof{table}{List of context item \& modifier, along with the direction of change, which caused to the largest magnitude of change, \(\Delta_{cm}\), for each personality trait.}
\label{tab:itemmaxdiffs}
\end{center}
\end{table}

To calculate the model scores, $X_{cm}$, for each trait, the models are evaluated on all ten items assigned to the trait, with each item serving as context once. This is done for each of the five modifiers, resulting in 10 (context items per trait) $\times$ 5 (modifiers per context item) $\times$ 10 (questionnaire items to be answered by the model) = 500 responses per trait and 10 (context items per trait) $\times$ 5 (modifiers per context item) = 50 scores (\(X_{cm}\)) per trait (one for each context). Context/modifier ratings (\(r_{cm}\)) are calculated to quantify the models' expected behavior in response to context. First, each modifier is assigned a modifier rating between -2 and 2 with -2 = \textit{never}, -1 = \textit{rarely}, 0 = \textit{sometimes}, 1 = \textit{often} and 2 = \textit{always}. Context items are given a context rating of -1 if the item negatively affected the trait score based on the IPIP scoring procedure, and 1 otherwise. The context ratings are multiplied by the modifier ratings to get the \(r_{cm}\). This value represents the expected relative change in trait score (expected behavior) when the corresponding context/modifier pair was used as context.

\begin{figure}
\centering
\includegraphics[width=0.7\columnwidth]{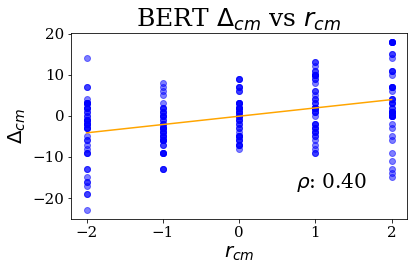}
\includegraphics[width=0.7\columnwidth]{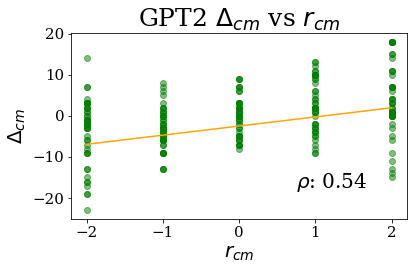}
\caption{\(\Delta_{cm}\) vs \(r_{cm}\) plots for data from all traits. We observe a consistent change in personality scores (\(\Delta_{cm}\)) across context items as the strength of quantifiers change. 
}
\label{fig:itemall}
\end{figure}

Next, the differences, $\Delta_{cm}$, between \(X_{cm}\) and \(X_{base}\) values are calculated and the correlation (Pearson\footnote{Pearson correlation coefficients used for all correlation measurements}) with the \(r_{cm}\) ratings measured (see Figure \ref{tab:itemmaxdiffs} for the context/modifier pairs with the largest $\Delta_{cm}$). One would expect \(X_{cm}\) evaluated on more positive \(r_{cm}\) to increase relative to \(X_{base}\) and vice versa. 
This is what we observe in Figure~\ref{fig:itemall}, where 
both BERT and GPT2 show significant correlations (0.40 and 0.54) between \(\Delta_{cm}\) and \(r_{cm}\) ($p < 0.01$, t-test). 
 
\begin{figure*}
\centering
\includegraphics[width=0.19\textwidth]{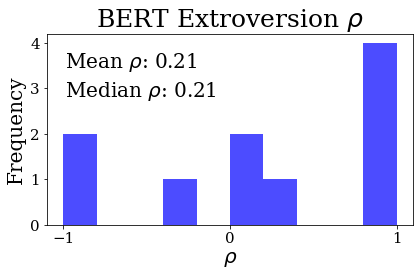}
\includegraphics[width=0.19\textwidth]{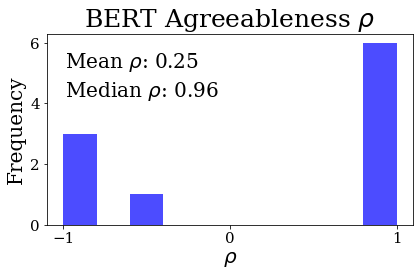}
\includegraphics[width=0.19\textwidth]{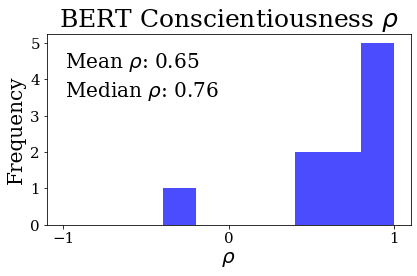}
\includegraphics[width=0.19\textwidth]{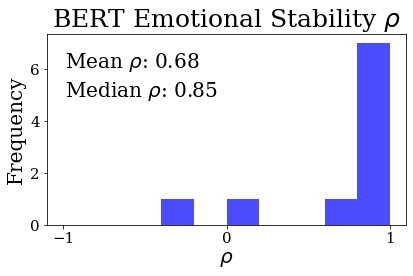}
\includegraphics[width=0.19\textwidth]{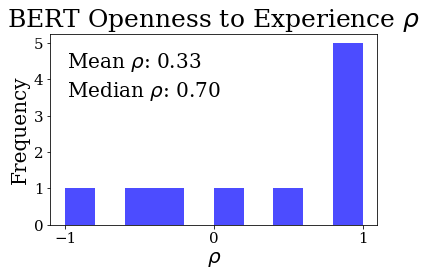}
\includegraphics[width=0.19\textwidth]{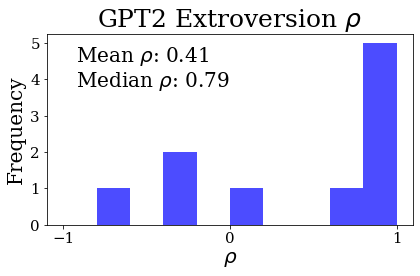}
\includegraphics[width=0.19\textwidth]{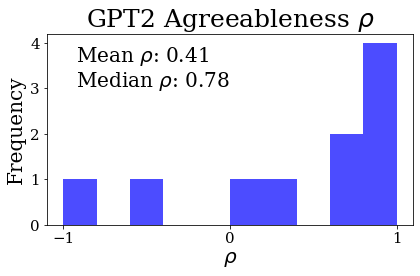}
\includegraphics[width=0.19\textwidth]{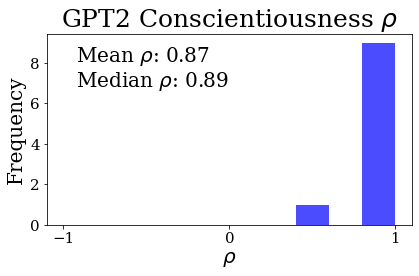}
\includegraphics[width=0.19\textwidth]{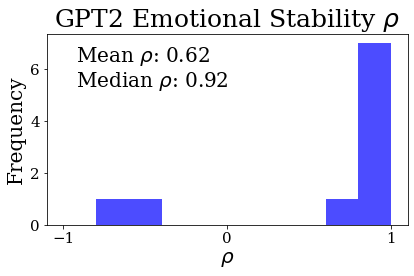}
\includegraphics[width=0.19\textwidth]{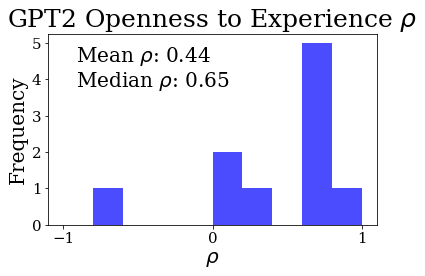}
\caption{Histograms of \(\rho\) by trait for \(\Delta_{cm}\) vs \(r_{cm}\) context item plots. Across all ten scenarios, a plurality of context items show a strong correlation (peak close to 1) between observed changes in personality traits and strengths of quantifiers in the context items.
}
\label{fig:itemdimcc}
\end{figure*}

Further, to look at the influence of individual context items as the strength of the modifier changes, we compute the correlation, \(\rho\), between \(\Delta_{cm}\) and \(r_{cm}\) for individual context items (correlation computed from 5 data points per context item, one for each modifier). Table \ref{tab:itemallcc} reports the mean and median values of these correlations. These results indicate a strong relationship between \(\Delta_{cm}\) and \(r_{cm}\). 
The mean values are significantly less than the medians, suggesting a left skew. For further analysis, the data was broken down by trait. The histograms in Figure \ref{fig:itemdimcc} depict \(\rho\) by trait and include summary statistics for this data.
    
\begin{table}
\small
\begin{center} 
\begin{tabular}{lrr}
\hline \cellcolor{gray!20}& \cellcolor{gray!20}\textbf{BERT} & \cellcolor{gray!20}\textbf{GPT2} \\ \hline
\textbf{Mean \(\rho\)} & 0.40 & 0.54\\
\textbf{Med \(\rho\)} & 0.84 & 0.81 \\
\hline
\end{tabular}
\end{center}
\captionof{table}{Mean \& median \(\rho\) from \(\Delta_{cm}\) vs \(r_{cm}\) plots by context item}
\label{tab:itemallcc}
\end{table}



Mean and median \(\rho\) from Figure \ref{fig:itemdimcc} plots suggest a positive linear correlation between \(\Delta_{cm}\) and \(r_{cm}\) amongst context item plots, with \textit{conscientiousness and emotional stability} having the strongest correlation for both BERT and GPT2. Groupings of \(\rho\) around 1 in \textit{conscientiousness and emotional stability} plots from Figure \ref{fig:itemdimcc} demonstrate this correlation. 

 GPT2 \textit{extroversion}, BERT \& GPT2 \textit{agreeableness} and BERT \textit{openness to experience} were subject to larger left skews and lower mean \(\rho\); their respective histograms show heavier groupings of \(\rho\) further left of the median. While BERT \textit{extroversion} didn't have a clear skew, it did have the lowest mean and median \(\rho\). It is possible that the effect of the five modifiers on \(\Delta_{cm}\) for a specific context item, such as BERT \textit{extroversion}, may follow a non-linear trend, resulting in lower correlations. 

A possible explanation for the larger skew in GPT2 \textit{extroversion}, BERT \& GPT2 \textit{agreeableness} and BERT \textit{openness to experience} histograms is that 
models may have had difficulty distinguishing between the double negative statements created by some context/modifier pairs (i.e. item 36 with modifier \textit{never}: ``I \textit{never} don't like to draw attention to myself."). This may have caused \(\Delta_{cm}\) to be negatively correlated with \(r_{cm}\), leading to an accumulation of \(\rho\) values near -1 for those traits.

Table~\ref{tab:itemmaxdiffs} shows the contexts that lead to the largest change for each of the personality traits for BERT and GPT2. We observe that all 10 contexts consist of the high-polarity quantifiers (either \textit{always} or \textit{never}), which consistent with the correlation results. Further, we note that for four of the five traits, the item context that leads to the largest change is common for the two models.

It is important to note a possible weakness with our approach of using questionnaire items as context. Since our evaluation also includes the same item during scoring, a language model could achieve a spurious correlation simply by copying the modifier choice mentioned in the context item. We experimented with adjustments\footnote{For the adjustments, we replaced the model response for question corresponding to the context item with the base model's response for the assessment. This means that the concerning context item can no longer contribute to $\Delta$. However, this also means that numbers with this adjustment cannot be directly compared with those without the since there are fewer sources of variation.} that would account for this issue and saw similar trends, with slightly lower but consistent correlation numbers (mean correlations of 0.25 and 0.40, compared with 0.40 and 0.54 for BERT and GPT2, statistically significant at $p < 0.05$, t-test).

\subsection{Analysis With Reddit Context}
\label{sec:reddit}
Next, we qualitatively analyze how personality traits of language models react to user-specific contexts. To acquire such context data, we curated data from Reddit threads asking individuals about their personality (see Appendix \ref{app:b} for a list of sources). 1119 responses were collected, the majority of which were in first person. Table~\ref{tab:redditcontext} shows two examples. Because GPT2 \& BERT tokenizers can’t accept more than 512 tokens, responses longer than this were truncated. The models were evaluated on the `Big Five' assessment (\S \ref{sec:design}) using each of the 1119 responses as context (Reddit context). For each Reddit context, scores, \(X_{reddit}\), were calculated for all 5 traits. The difference between \(X_{reddit}\) and \(X_{base}\) was calculated as $\Delta_{reddit}$. 

\begin{table}
\small
\centering
\begin{tabular}{l}
\hline 
\multicolumn{1}{c}{\cellcolor{gray!20}\textbf{Context}} \\ \hline
Subdued until I really get to know someone. \\
\hline
I am polite but not friendly. I do not feel the need\\
to hang around with others and spend most of my time\\ 
reading, listening  to music, gaming or watching films.\\
Getting to know me well is quite a  challenge I suppose,\\
but my few friends and I  have a lot of fun when we\\
meet (usually at university or online, rarely elsewhere\\
irl). I'd say I am patient, rational and a guy with a\\
big heart for the ones I care for. 
\\
\hline
\end{tabular}
\caption{Examples of Reddit data context.}
\label{tab:redditcontext}
\end{table}


To interpret what words or phrases in the contexts affect the language models' personality traits, we train regression models on bag-of-words and n-gram (with $n=2$ and $n=3$) representations of the Reddit contexts as input,  and $\Delta_{reddit}$ values as labels. Since the 
goal was to analyze attributes in the contexts that caused substantial shifts in trait scores, we only consider contexts with \(\|\Delta_{reddit}\|\ge1\). 
Next, we extracted the ten most positive and most negative feature weights for each trait, and performed a qualitative analysis of these features. 
We note that for \textit{extroversion}, phrases such as `friendly', `great' and `no problem' are among the highest positively weighted phrases, whereas phrases such as `stubborn' and `don't like people' are among the most negatively weighted. For \textit{agreeableness}, phrases like `love' and `loyal' are positively weighted, whereas phrases such as `lazy', `asshole' and expletives are weighted highly negative.
On the whole, our qualitative analysis revealed that the changes in personality scores for most traits conformed with a human understanding of the most highly weighted features. 
As further examples, phrases such as `hang out with' caused a positive shift in trait score for \textit{openness to experience}, while `lack of motivation' causes a negative shift for \textit{conscientiousness}. However, some other strongly weighted phrases appeared to have little relation to the trait definition or expected connotation. 
There were fewer phrases for GPT2 \textit{openness to experience}, GPT2 negatively weighted \textit{agreeableness}, and GPT2 negatively weighted \textit{extroversion} that caused shifts in the expected direction.
This was consistent with results from \textit{\S \ref{sec:item}}, where these traits exhibited the weakest relative positive correlations. 
Appendix~\ref{app:b} contains the full lists of highly weighted features for each trait.


\begin{table}[h]
\small
\centering
\begin{tabular}{l}
\hline
\multicolumn{1}{c}{\textbf{Context}}  \\
\hline
\multicolumn{1}{c}{\cellcolor{gray!20}\textbf{\textit{Undirected Response}}} \\
\hline
I am a very open-minded, polite person and always crave\\
new experiences. At work I manage a team of software\\
developers and we often have to come up with new ideas.\\
I went to college and majored in computer science, and\\
enjoyed the experience.  I have met many like-minded\\
people and I enjoy speaking with them about a lot of \\
various topics. I am sometimes shy around people who I\\
don't know well, but I try to be welcoming and warm to\\
everyone I meet. I try to do something fun every week,\\
even if I'm quite busy, like having a BBQ or watching a\\
movie. I have a wife whom I love and we live together in\\
a nice single-family home. \\
\hline
\multicolumn{1}{c}{\cellcolor{gray!20}\textbf{\textit{Directed Response}}} \\
\hline
I consider myself to be someone that is quiet and\\
reserved. I do not like to talk that much unless I have\\
to. I am fine with being by myself and enjoying the peace\\ and quiet. I usually agree with people more often than\\
not. I am a polite and kind person. I am mostly honest,\\
but I will lie if I feel it is necessary or if it benefits me\\
in a huge way. I am easily irritated by things and I have \\ anxiety issues. I like to be open minded and learn about\\
new things. I am a curious person. I enjoy having a plan \\
and following it.
\\
\hline
\end{tabular}
\caption{Examples of survey data contexts.}
\label{tab:surveycontext}
\end{table}

\subsection{Analysis With Psychometric Survey Data}
\label{sec:survey}
The previous sections indicate that language models can pick up on personality traits from context. This suggests the following question: can these models be used to estimate an individual's personality? In theory, this would be done by evaluating on the `Big Five' personality assessment using context describing the individual. This can aid in personality characterization in cases where it is not feasible for a subject to manually undergo a personality assessment. We investigate this through the following experiment. 

Using Amazon Mechanical Turk, subjects were asked to complete the 50-item `Big Five' personality assessment outlined in \textit{\S \ref{sec:design}} (the assessment was not modified to a sentence completion format as was done for model testing) and provide a 75-150 word description of their personality (see Appendix \ref{app:c} for survey instructions). Responses were manually filtered and low effort attempts discarded, resulting in 404 retained responses. Two variations of the study were adopted: the subjects for 199 of the responses were provided a brief summary of the `Big Five' personality traits and asked to consider, but not specifically reference, these traits in their descriptions. We refer to these responses as the \textit{Directed Responses} data set. The remaining 205 subjects were not provided with this summary and their responses make up the \textit{Undirected Responses} data set. Table \ref{tab:surveycontext} shows examples of collected descriptions. Despite asking for personality descriptions upwards of 75 words, around a fourth of the responses fell below this limit. The concern was that data with low word counts may not have enough context. Thus, we experiment with filtering the responses by removing outliers (based on the interquartile ranges) and including minimum thresholds on the description length (75 and 100). 

\begin{figure}
\centering
\includegraphics[width=.7\columnwidth]{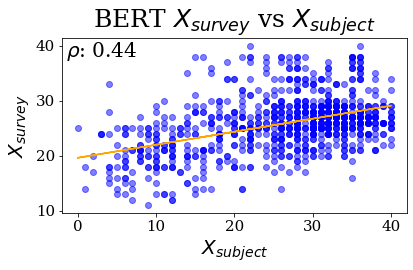}
\includegraphics[width=.7\columnwidth]{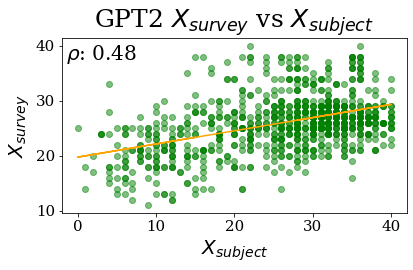}
\caption{BERT \& GPT2 \(X_{survey}\) vs \(X_{subject}\) plots (\textit{Directed Responses} with outliers removed). Regression lines and correlation coefficients (\(\rho\)) are shown.}
\label{fig:surveyall}
\end{figure}

Human subject scores, \(X_{subject}\), were calculated for each assessment, using the same scoring procedure as previously described in \S \ref{sec:design}. The models were subsequently evaluated on the `Big Five' personality assessment using the subjects' personality descriptions as context, yielding \(X_{survey}\) scores corresponding to each subject. Figure \ref{fig:surveyall} shows a plot of \(X_{survey}\) against \(X_{subject}\) for individual subjects, and indicates strong correlations (0.48 for GPT2 and 0.44 for BERT) between predicted personality traits of human subjects based on their personality descriptions (\textit{Directed Responses}) and their actual psychometric assessment scores. Table~\ref{tab:surveyalld2} shows a summary of the correlation statistics for the two different data sets and different filters. We note that there are only marginal differences in correlations between the two datasets, inspite of their different characteristics. 
While more specific testing is required to determine causal factors that explain these observed correlation values, they suggest the potential for using language models as probes for personality traits in free text. 

\begin{figure}
\centering
\includegraphics[width=0.8 \columnwidth]{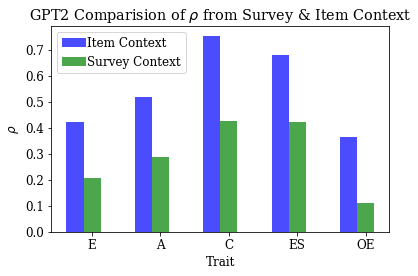}
\caption{The plot compares \(\rho\) from model evaluation with item context (\textit{\S \ref{sec:item}}) and survey context (\textit{\S \ref{sec:survey}}). Survey context \(\rho\) shown here are from \textit{Undirected Responses} (\textit{c}\(\geq\) 100). In both cases, $\rho$ measures the correlation between trait scores with context and expected behavior. The variables used to quantify expected behavior differ between experiments.
}
\label{fig:cccomp}
\end{figure}

\begin{table}
\small
\centering
\begin{tabular}{lrrr}
\hline \textbf{Trait} & \textbf{\(\rho_{no-outlier}\)} & \textbf{\(\rho_{c \geq 75}\)} &  \textbf{\(\rho_{c \geq 100}\)} \\ \hline
\multicolumn{4}{c}{\cellcolor{gray!20}\textbf{\textit{Undirected Responses}}} \\
\hline
BERT & 0.40 & 0.39 & 0.41 \\
GPT2 & 0.48 & 0.43 & 0.48 \\
\hline
\multicolumn{4}{c}{\cellcolor{gray!20}\textbf{\textit{Directed Responses}}} \\
\hline
BERT & 0.44 & 0.42 & 0.39 \\
GPT2 & 0.48 & 0.43 & 0.42 \\
\hline
\end{tabular}
\caption{\(\rho\) for \(X_{survey}\) vs \(X_{subject}\) for data filtered by removing outliers and enforcing word counts.}
\label{tab:surveyalld2}
\end{table}

Figure \ref{fig:cccomp} plots the correlations \(\rho\) (outliers removed) for the individual personality traits, and also includes correlation coefficients from \textit{\S \ref{sec:item}}. While the correlations from both sections are measured for different variables, they both represent a general relationship between observed personality traits of language models and the expected behavior (from two different types of contexts). While there are positive correlations for all ten scenarios, correlations from survey contexts are smaller than those from item contexts. This is not surprising since item contexts are specifically handpicked by domain experts to be relevant to specific personality traits, while survey contexts are open-ended. 
These promising results come despite the data containing some low-effort responses, which might fail to adequately express subject personalities. 

\subsection{Observed Ranges of Personality Traits}
\begin{figure}
\centering
\includegraphics[width=\columnwidth]{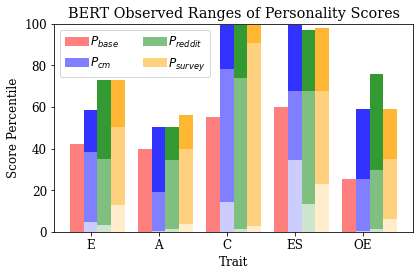}
\caption{Chart showing observed ranges of personality traits (in terms of human percentiles) exhibited by BERT,  when conditioned on different context types. These include scores from the base model (\(P_{base}\)) and ranges of scores from the three context types: item (\(P_{cm}\)), Reddit (\(P_{reddit}\)) and survey (\(P_{survey}\)). Bars for context-based scores show the percentile of the minimum, median, and maximum-scoring context, in ascending order. The lightest shade of each color indicates the minimum, the darkest indicates the maximum and the intermediate shade indicates the median.
}
\label{fig:itembar}
\end{figure}

In the previous subsections, we investigated priming language models with different types of contexts to manipulate their personality traits. Figure \ref{fig:itembar} visually summarizes and compares the observed ranges of personality trait scores for different contexts, grouped by context types. The four columns for each trait represent the scores achieved by the base model (no context), and the ranges of scores achieved by the different types of contexts. The minimum, median and maximum scores for each context type are indicated by different shades on each bar. We observe that the different contexts lead to a remarkable range of scores for all five personality traits. In particular, we note that for two of the traits (\textit{conscientiousness} and \textit{emotional stability}), the models actually achieve the full range of human scores (nearly 0 to 100 percentile). Curiously, for all five traits, different contexts are able to achieve very low scores (\(< 10\) percentile). However, the models particularly struggle with achieving high scores for \textit{agreeableness}. For all traits, the contexts lead to a substantial range of behaviors compared with the base model scores. On the other hand, there are no major differences in the ranges of scores observable using the three types of contexts that we experiment with.  

\section{Gender Differences in Personality}
\label{sec:gender}
Previous research has found differences in personality with respect to attributes such as 
age and gender~\citep{Srivastava:03}. We explored whether language models are also influenced by such attributes. For this, the assessment from \textit{\S \ref{sec:design}} was modified to use names. 
For example, for the name David, the item ``I \{blank\} \textit{have} excellent ideas" was changed to ``David \{blank\} \textit{has} excellent ideas". BERT \& GPT2 were evaluated on this modified assessment for the 20 most common male and female names in the US over the last 100 years, according to the US Social Security Administration, resulting in 20 \(X_{male}\) and 20 \(X_{female}\) scores. Table~\ref{tab:gender} shows the means of these scores and the corresponding human population percentiles (\(P_{male}\), \(P_{female}\)). 
We note that mean female scores are higher for \textit{agreeableness}, \textit{conscientiousness}, and \textit{emotional stability} for both BERT and GPT2. In fact, mean male scores are only higher for GPT2's \textit{extroversion} and \textit{openness to experience}. While the sample sizes for these results are insufficient to make significant inferences, they agree with psychological research on higher levels of \textit{agreeableness} and \textit{conscientiousness} in women. On the other hand, literature suggests that women show lower levels of \textit{emotional stability} (higher neuroticism), which diverges from model predictions. 



\begin{table}
\small
\centering
\begin{tabular}{lrr}
\hline \textbf{Trait} & \textbf{\(P_{male}\) (\%)} & \textbf{\(P_{female}\) (\%)} \\ \hline
\multicolumn{3}{c}{\cellcolor{gray!20}\textbf{BERT}} \\
\hline 
E & 31 & 35 \\
A & 29 & 34 \\
C & 49 & 68 \\
ES & 47 & 56 \\
OE & 39 & 39 \\
\hline
\multicolumn{3}{c}{\cellcolor{gray!20}\textbf{GPT2}} \\
\hline 
E & 50 & 42\\
A & 16 & 19 \\
C & 59 & 64 \\
ES & 35 & 43 \\
OE & 33 & 28 \\
\hline
\end{tabular}
\caption{Human population percentile for mean \(X_{male}\) (\(P_{male}\)) and mean \(X_{female}\) (\(P_{female}\)).}
\label{tab:gender}
\end{table}

\section{Discussion}
\label{sec:discuss}
We have presented a simple and effective approach for controlling the personality traits of language models. Further, we show that 
such models could also be used with language-based question answering to predict personality traits of human users, enabling assessment in cases where quality participation is difficult to attain. 
Future work can explore the use of alternate personality personality taxonomies. Similarly, there is a large and ever-growing variety of language models apart from BERT and GPT2. It is unclear to what extent our findings would generalize to other language models, particularly those such as GPT3~\cite{brown2020language} and MT-NLG~\cite{smith2022using} with a significantly larger number of parameters. Finally, the role that pretraining data plays on personality traits is an important question for future exploration. 

\section*{Ethics and Broader Impact} 
The `Big Five' assessment items and scoring procedure were drawn from free public resources and open source implementations of BERT, GPT2 and the logistic regression classifier were used \citep{huggingface:22, scikit1:22}. Reddit data was scraped from public threads and no usernames or other identifiable markers were collated. The crowd-sourced survey data was collected using Amazon Mechanical Turk (AMT), with the permission of all participants. No personally identifiable markers were stored and participants were compensated fairly, with a payment rate (\$2.00/task w/ est. completion time of 15 min) significantly greater than AMT averages \citep{adams:2018}. Participants were also informed that the data would be used for academic purposes.

The overarching goal of this line of research is to investigate aspects of personality in language models, which are increasingly being used in a number of NLP applications. Since AI systems that use these technologies are growing ever pervasive, and as humans tend to anthropomorphize such systems (such as Siri and Alexa), understanding and controlling their personalities can have both broad and deep consequences. This is especially true for applications in domains such as education and mental health, where interactions with these systems can have lasting personal impacts on their users. 

Finally, if the personalities of AI systems can be manipulated in the ways that our research suggests, there is a serious risk of such systems being manipulated to be hostile or disagreeable to their users through specific attacks. Developing methods through which language models could be made immune to such attacks would then be a necessary consideration before fielding such systems.
\bibliography{custom}

\clearpage
\begin{appendices}

\section{Model Background}
\label{app:z}


BERT, which stands for Bidirectional Encoder Representations from Transformers, is a transformer-based deep learning model for natural language processing \citep{Devlin:19}. The model is pre-trained on unlabeled data from the 800M word BooksCorpus and 2500M word English Wikipedia corpora. While BERT can be fine-tuned for autoregressive language modeling tasks, it is pretrained for masked language modeling. This study uses a BERT model from HuggingFaces’s Transformer Python Library with a language model head for masked language modeling. No fine-tuning was done to the model. GPT2, which stands for Generative Pre-trained Transformer 2, is a general-purpose learning transformer model developed by OpenAI in 2018 \citep{Radford:22}. Like BERT, this model is also pretrained on unlabeled data from the 800M word BooksCorpus. The study used Hugginface’s GPT2 model with a language model head for autoregressive language modeling. As with BERT, no fine-tuning took place.

\renewcommand{\thetable}{\Alph{section}.\arabic{table}}
\renewcommand{\thetable}{\Alph{section}.\arabic{figure}}

\begin{figure}[t]
\section{Experiment Design Items}
\label{app:a}
\smallskip
\centering
\includegraphics[width=.89\columnwidth]{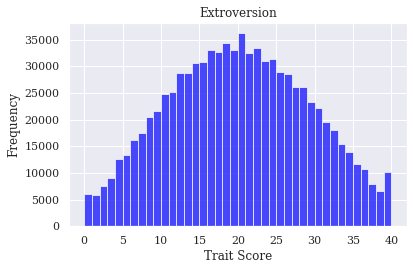}
\includegraphics[width=.89\columnwidth]{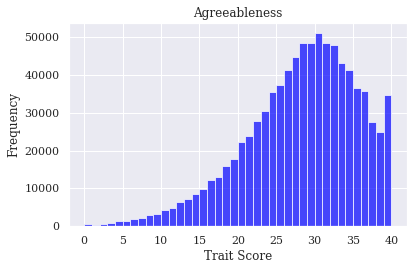}
\includegraphics[width=.89\columnwidth]{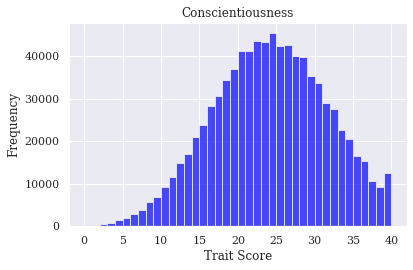}
\includegraphics[width=.89\columnwidth]{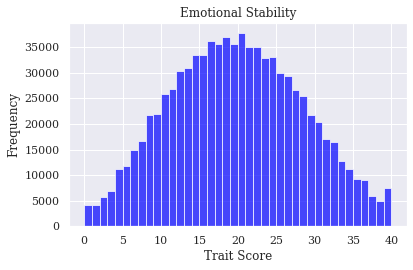}
\includegraphics[width=.89\columnwidth]{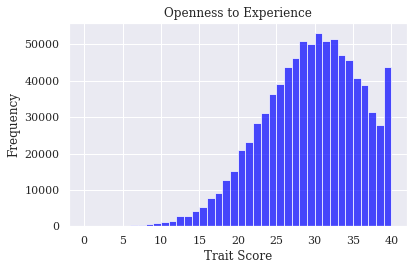}
\caption{Human distributions of `Big Five' trait scores.}
\label{fig:humandist}
\end{figure}

\setcounter{table}{0}

\begin{table*}
\small
\centering
\begin{tabular}{lc}
\hline
\cellcolor{gray!20}\textbf{Item} & \cellcolor{gray!20}\textbf{Associated Trait }\\
\hline 
I am \{blank\} the life of the party. &E \\
I \{blank\} feel little concern for others. &A \\
I am \{blank\} prepared. &C \\
I \{blank\} get stressed out easily. &ES \\
I \{blank\} have a rich vocabulary. &OE \\
I \{blank\} don't talk a lot. &E \\
I am \{blank\} interested in people. &A \\
I \{blank\} leave my belongings around. &C \\
I am \{blank\} relaxed most of the time. &ES \\
I \{blank\} have difficulty understanding abstract ideas. &OE \\
I \{blank\} feel comfortable around people. &E \\
I \{blank\} insult people. &A \\
I \{blank\} pay attention to details. &C \\
I \{blank\} worry about things. &ES \\
I \{blank\} have a vivid imagination. &OE \\
I \{blank\} keep in the background. &E \\
I \{blank\} sympathize with others' feelings. &A \\
I \{blank\} make a mess of things. &C \\
I \{blank\} seldom feel blue. &ES \\
I am \{blank\} not interested in abstract ideas. &OE \\
I \{blank\} start conversations. &E\\
I am \{blank\} not interested in other people's problems. &A\\
I \{blank\} get chores done right away. &C \\ 
I am \{blank\} easily disturbed. &ES \\
I \{blank\} have excellent ideas. &OE \\
I \{blank\} have little to say. &E \\
I \{blank\} have a soft heart. &A \\
I \{blank\} forget to put things back in their proper place. &C \\
I \{blank\} get upset easily. &ES \\
I \{blank\} do not have a good imagination. &OE \\
I \{blank\} talk to a lot of different people at parties. &E \\
I am \{blank\}    not really interested in others. &A \\
I \{blank\} like order. &C \\
I \{blank\} change my mood a lot. &ES \\
I am \{blank\} quick to understand things. &OE \\
I \{blank\} don't like to draw attention to myself. &E \\
I \{blank\} take time out for others. &A \\
I \{blank\} shirk my duties. &C \\
I \{blank\} have frequent mood swings. &ES \\
I \{blank\} use difficult words. &OE \\ 
I \{blank\} don't mind being the center of attention. &E \\
I \{blank\} feel others' emotions. &A \\
I \{blank\} follow a schedule. &C \\
I \{blank\} get irritated easily. &ES \\
I \{blank\} spend time reflecting on things. &OE \\
I am \{blank\} quiet around strangers. &E \\
I \{blank\} make people feel at ease. &A \\ 
I am \{blank\} exacting in my work. &C \\ 
I \{blank\} feel blue. &ES \\
I am \{blank\} full of ideas. &OE \\
\hline
\end{tabular}
\caption{Adjusted `Big Five' Personality Assessment Items.}
\label{tab:a1}
\end{table*}

\begin{table*}
\small
\centering
\begin{tabular}{lrrr}
\hline\cellcolor{gray!20} \textbf{Trait} &
\cellcolor{gray!20}\textbf{Median} &
\cellcolor{gray!20}\textbf{Mean ($\mu$)} &  \cellcolor{gray!20}\textbf{SD ($\sigma$)} \\ \hline
E & 20 & 19.60  &  9.10 \\
A & 29 & 27.74 & 7.29 \\
C & 24 & 23.66 & 7.37\\
ES & 19 & 19.33 & 8.59 \\
OE & 29 & 28.99 & 6.30 \\
\hline
\end{tabular}
\caption{Human Population Distribution of `Big Five' Personality Traits.}
\label{tab:a3}
\end{table*}

\begin{table*}
\small
\centering
\begin{tabular}{lrrr}
\hline \cellcolor{gray!20}\textbf{Trait} & \cellcolor{gray!20}\textbf{Base Value} & \cellcolor{gray!20}\textbf{Positively Scored Item \#} & \cellcolor{gray!20}\textbf{Negatively Scored Item \# } \\ \hline
E & 20 & 1, 11, 21, 31, 41 &  6, 16, 26, 36, 46 \\
A & 14 & 7, 17, 27, 37, 42, 47 & 2, 12, 22, 32 \\
C & 14 & 3, 13, 23, 33, 43, 48 & 8, 18, 28, 38\\
ES & 38 & 9, 19 & 4, 14, 24, 29, 34, 39, 44, 49 \\
OE & 8 & 5, 15, 25, 35, 40, 45, 50  & 10, 20, 30 \\
\hline
\end{tabular}
\caption{`Big Five' Personality Item Scoring Procedure.}
\label{tab:a2}
\end{table*}

\setcounter{table}{0}

\setcounter{table}{0}

\begin{table*}
\small
\section{Item Context Evaluation Tables}
\smallskip
\label{app:y}
\small
\centering 
\begin{tabular}{rrrrr}
\hline \textbf{\(r_{cm}\)} & \textbf{Mean  \(\Delta_{cm}\)} & \textbf{Med  \(\Delta_{cm}\)} & \textbf{\(\Delta_{cm}\) SD} & \textbf{Confidence Interval}\\ \hline
\multicolumn{5}{c}{\cellcolor{gray!20}\textbf{BERT}} \\
\hline 
-2 & -3.36 & -2.0 &  7.49 & [-5.51, -1.21] \\
-1 & -3.18 & -3.50 & 4.81 & [-4.56, -1.80] \\
0 & -0.02 & 0.00 & 4.51 & [-1.32, 1.28] \\
1 & 2.42 & 2.00 & 6.17 & [0.648, 4.19]\\
2 & 3.96  & 3.00 &  8.33 & [1.57, 6.35]\\
\hline
\multicolumn{5}{c}{\cellcolor{gray!20}\textbf{GPT2}} \\
\hline 
-2 & -7.34 & -8.0 &  6.38 & [-9.17, -5.51] \\
-1 & -4.58 & -4.0 & 4.32 & [-5.82, -3.34] \\
0 & -2.06 & -1.0 & 4.24 & [-3.28, -0.84] \\
1 & 0.0 & 0.0 & 3.13 & [-0.90, 0.90]\\
2 & 1.56  & 1.0 &  5.78 & [-0.10, 3.22]\\
\hline
\end{tabular}
\caption{Statistics from \(\Delta_{cm}\) vs \(r_{cm}\) plots containing data from all traits. Statistics include mean, median, standard deviation and a confidence interval for \(\Delta_{cm}\) at each \(r_{cm}\).
}
\label{tab:y1}
\end{table*}

\setcounter{table}{0}
\begin{table*}
\small
\section{Reddit Context Evaluation Tables}
\smallskip
\label{app:b}
\centering
\begin{tabular}{l}
\hline
\multicolumn{1}{c}{\cellcolor{gray!20}\textbf{Reddit Context Sources}} \\ 
\hline
reddit.com/r/AskReddit/comments/k3dhnt/how\_would\_you\_describe\_your\_personality/ \\ 
reddit.com/r/AskReddit/comments/q4ga1j/redditors\_what\_is\_your\_personality/ \\
reddit.com/r/AskReddit/comments/68jl8g/how\_can\_you\_describe\_your\_personality/ \\ 
reddit.com/r/AskReddit/comments/ayjgyz/whats\_your\_personality\_like/ \\
reddit.com/r/AskReddit/comments/9xjahw/how\_would\_you\_describe\_your\_personality/ \\
reddit.com/r/AskWomen/comments/c1gr4a/how\_would\_you\_describe\_your\_personality/ \\
reddit.com/r/AskWomen/comments/7x23zg/what\_are\_your\_most\_defining\_personalitycharacter/ \\
reddit.com/r/CasualConversation/comments/5xtckg/how\_would\_you\_describe\_your\_personality/ \\
reddit.com/r/AskReddit/comments/aewroe/how\_would\_you\_describe\_your\_personality/ \\
reddit.com/r/AskMen/comments/c0grgv/how\_would\_you\_describe\_your\_personality/ \\
reddit.com/r/AskReddit/comments/pzm3in/how\_would\_you\_describe\_your\_personality/ \\
reddit.com/r/AskReddit/comments/bem0ro/how\_would\_you\_describe\_your\_personality/ \\
reddit.com/r/AskReddit/comments/1w9yp0/what\_is\_your\_best\_personality\_trait/ \\
reddit.com/r/AskReddit/comments/a499ng/what\_is\_your\_worst\_personality\_trait/ \\
reddit.com/r/AskReddit/comments/6onwek/what\_is\_your\_worst\_personality\_trait/ \\
reddit.com/r/AskReddit/comments/2d7l2i/serious\_reddit\_what\_is\_your\_worst\_character\_trait/ \\
reddit.com/r/AskReddit/comments/449cu7/serious\_how\_would\_you\_describe\_your\_personality/ \\ 
\hline
\end{tabular}
\caption{Domain names of threads that were scraped to collect Reddit context.}
\label{tab:b3}
\end{table*}

\begin{table*}
\small
\centering 
\begin{tabular}{lrrrrr}
\hline \textbf{Trait} & \textbf{Mean $\Delta_{reddit}$} & \textbf{Med $\Delta_{reddit}$} & \textbf{$\Delta_{reddit}$ SD} & \textbf{5 Max $\Delta_{reddit}$} & \textbf{5 Min $\Delta_{reddit}$} \\ \hline
\multicolumn{6}{c}{\cellcolor{gray!20}\textbf{BERT}} \\
\hline 
E & -2.28 & -2 & 4.04 & 8, 7, 7, 6, 5 & -14, -13, -13, -13, -13 \\
A & -2.02 & -1 & 3.38 & 2, 2, 2, 2, 2 & -19, -18, -15, -15, -15\\
C & 3.77 & 4 & 5.17 & 15, 15, 15, 15, 13 & -17, -17, -16, -14, -13\\
ES & 1.71 & 2 & 2.29 & 14, 14, 13, 13, 12 & -12, -10, -10, -10, -10\\
OE & 1.74 & 1 & 2.17 & 9, 7, 7, 7, 7 & -11, -11, -8, -8, -7\\
\hline
\multicolumn{6}{c}{\cellcolor{gray!20}\textbf{GPT2}} \\
\hline 
E & -3.73 & -4 &  3.33 & 7, 5, 5, 4, 4 & -14, -10, -10, -10, -10 \\
A &-0.98 & -1 & 4.26 & 13, 10, 8, 7, 7 & -17, -15, -15, -15, -14 \\
C & -0.27 & 0 & 4.27 & 11, 11, 11, 11, 9 & -20, -16, -16, -16, -15 \\
ES & -3.83 & -3 & 6.27 & 8, 8, 8, 8, 8 & -21, -21, -21, -21, -21 \\
OE & -1.91 & -2 & 3.21 & 4, 4, 4, 4, 4 & -15, -12, -12, -12, -12 \\
\hline
\end{tabular}
\caption{$\Delta_{reddit}$ summary statistics. Statistics include mean, median and standard deviation, as well as 5 largest and 5 smallest $\Delta_{reddit}$.}
\label{tab:b4}
\end{table*}

\begin{table*}
\small
\centering
\begin{tabular}{c}
\textbf{BERT} \\
\noindent\fbox{%
    \parbox{\textwidth}{%
\textit{Extroversion}
\begin{itemize}
\item Notable Positively Weighted Phrases: 'friendly',  'great', 'good', 'quite', 'laugh', 'please', 'sense of', 'thanks for', 'really good', 'and friendly', 'no problem', 'to please', 'my sense of', 'finish everything start', 'enthusiastic but sensitive'
\end{itemize}
\begin{itemize}
\item Notable Negatively Weighted Phrases: 'question', 'stubborn', 'why', 'lack', 'fuck', 'fucking', 'hate', 'not', 'lack of', 'too much', 'don know', 'don like', 'too easily', 'way too', 'don like people', 'you go out', 'don know how', 'don['t] know what'
\end{itemize}
\textit{Agreeableness}
\begin{itemize}
\item Notable Positively Weighted Phrases: 'will', 'friendly', 'lol', 'love', 'loyal', 'calm', 'yup', 'does', 'honesty', 'laid back', 'go out', 'thanks for', 'really good', 'out with me', 'friendly polite and', 'really good listener', 'true to myself', 'my sense of'
\end{itemize}
\begin{itemize}
\item Notable Negatively Weighted Phrases: 'lack', 'didn['t]', 'won['t], 'lazy', 'fucking', 'self', 'worst', 'lack of', 'too easily', 'don like', 'the worst', 'being too', 'have no', 'don like people', 'lack of motivation', 'don know how', 'my worst trait', 'also my worst', 'too honest sometimes', 'doesn['t] talk much'
\end{itemize}
\textit{Conscientiousness}
\begin{itemize}
\item Notable Positively Weighted Phrases: 'am', 'friendly', 'just', 'calm', 'believe', 'can be', 'of people', 'tend to', 'feel like', 'the most humble', 'most humble person', 'my sense of', 'get to know', 'friendly polite and', 'get along with', 'people like me'
\end{itemize}
\begin{itemize}
\item Notable Negatively Weighted Phrases: 'lack', 'no', 'lazy', 'inability', 'fucks', 'half', 'lack of', 'fuck off', 'don like', 'inability to', 'don like people', 'you go out', 'lack of motivation', 'don even know', 'monotonous and impulsive'
\end{itemize}
\textit{Emotional Stability}
\begin{itemize}
\item Notable Positively Weighted Phrases: 'will', 'feel', 'out with me', 'go out with', 'will you go', 'the most humble'
\end{itemize}
\begin{itemize}
\item Notable Negatively Weighted Phrases: 'no', 'off', 'hypercritical', 'overthinking', 'lack of', 'easily distracted', 'doesn talk', 'don even', 'too easily distracted', 'lack of motivation', 'doesn talk much', 'don even know', 'unrelatable is strange', 'is strange one', 'this said foreskin'
\end{itemize}
\textit{Openness to Experience}
\begin{itemize}
\item Notable Positively Weighted Phrases: 'most', 'like', 'me to', 'out with', 'like me', 'like to', 'want to', 'with me', 'out with me', 'will you go', 'want to be', 'all the time', 'for me to', 'hang out with'
\end{itemize}
\begin{itemize}
\item Notable Negatively Weighted Phrases: 'lack', 'never', 'fucks', 'sad', 'nothing', 'lack empathy', 'the complainer', 'no confidence', 'lack of', 'easily distracted', 'blame helicopter', 'helicopter parents', 'never say sorry', 'blame helicopter parents', 'too easily distracted', 'finish projects after', 'never finish projects', 'procrastination out of', 'my lack of', 'lack of personality', 'too many fucks'
\end{itemize}
    }%
}
\end{tabular}
\caption{Analysis of highest weighted phrases from BERT logistic regression.}
\label{tab:b1}
\end{table*}

\begin{table*}
\small
\centering
\begin{tabular}{c}
\textbf{GPT2} \\
\noindent\fbox{%
    \parbox{\textwidth}{%

\textit{Extroversion}
\begin{itemize}
\item Notable Positively Weighted Phrases: 'believe', 'loyal', 'curious', 'best', 'passionate', 'enjoy', 'bright', 'hard working', 'no problem', 'am nice',  'my amazing modesty', 'smooth bright epic', 'patient and flexible', 'great with children', 'calm cool collected'
\end{itemize}
\begin{itemize}
\item Notable Negatively Weighted Phrases: 'introverted', 'lack of', 'laid back', 'don know how'
\end{itemize}
\textit{Agreeableness}
\begin{itemize}
\item Notable Positively Weighted Phrases: 'friendly', 'loyal', 'honest', 'gay', 'humor', 'like people', 'thanks for', 'to please', 'and friendly', 'no problem', 'friendly polite and', 'patient and flexible', 'calm cool collected', 'honesty being straightforward'
\end{itemize}
\begin{itemize}
\item Notable Negatively Weighted Phrases: 'too easily', 'too much', 'lack of', 'you go out', 'don know what', 'self', 'asshole'
\end{itemize}
\textit{Conscientiousness}
\begin{itemize}
\item Notable Positively Weighted Phrases: 'smile',  'thanks for', 'no problem', 'friendly polite and', 'really good listener', 'true to myself', 'patient and flexible'
\end{itemize}
\begin{itemize}
\item Notable Negatively Weighted Phrases: 'stop', 'jealousy', 'lazy', 'hate', 'lack', 'fuck', 'worst', 'lack of', 'too easily', 'fuck off', 'too nice', 'don know',  'don know how', 'lack of motivation', 'don even know', 'my worst trait', 'damn it uncle', 'depressed as shit'
\end{itemize}
\textit{Emotional Stability}
\begin{itemize}
\item Notable Positively Weighted Phrases: 'friendly', 'calm', 'easy', 'honesty', 'laid back', 'hard working', 'calm and', 'humble am', 'polite and', 'no problem', 'out with me', 'the most humble'
\end{itemize}
\begin{itemize}
\item Notable Negatively Weighted Phrases: 'lack', 'anxious', 'lazy', 'jealousy', 'lack of', 'don know', 'too easily', 'don like', 'don like people', 'don know how', 'lack of motivation', 'don even know'
\end{itemize}
\textit{Openness to Experience}
\begin{itemize}
\item Notable Positively Weighted Phrases: 'understand', 'having', 'wanting', 'thoughts', 'thanks for', 'too nice', 'no problem', 'can relate', 'being too nice', 'that just confidence'
\end{itemize}
\begin{itemize}
\item Notable Negatively Weighted Phrases: 'fuck', 'myself', 'cynical', 'lack', 'boring', 'lack of', 'don like people'
\end{itemize}
}
}
\end{tabular}
\caption{Analysis of highest weighted phrases from GPT2 logistic regression.}
\label{tab:b2}
\end{table*}

\setcounter{table}{0}
\begin{table*}
\section{Survey Context Evaluation Tables}
\smallskip

\label{app:c}
\small
\centering 
\begin{tabular}{l}
\hline 
\multicolumn{1}{c}{\cellcolor{gray!20}\textbf{Part 1 Instruction}} \\
\hline 
There are two parts to this questionnaire. In the first part (on this page), you will be shown 50 questions, \\
and need to choose a response which best matches your personality. In the second part (on the next page), \\
you will be asked to write a short (75-150 word) description of your personality in free text. Participants \\
will only be compensated if they respond to all questions. \\
\hline
\multicolumn{1}{c}{\cellcolor{gray!20}\textbf{Part 2 Instruction}} \\
\hline 
In between 75 and 150 words, please describe your personality [\textit{Directed responses}: as it relates to the 5 personality traits \\
outlined above. Be sure not to use the name of the personality traits themselves in your response]. \\
\end{tabular}
\caption{Data collection survey instructions.
}
\label{tab:c1}
\end{table*}

\end{appendices}

\end{document}